\documentclass[10pt,twocolumn,letterpaper]{article}

\usepackage{cvpr}              

\usepackage[dvipsnames]{xcolor}

\usepackage[normalem]{ulem}
\usepackage{multirow}
\usepackage{colortbl}
\usepackage{pifont}
\usepackage{bbding}
\def\halfcheckmark{\ding{52}\textsuperscript{\kern-0.50em\small\ding{55}}}

\newcommand{\cmark}{\ding{52}}%

\usepackage{algorithm}
\usepackage{algorithmic}

\definecolor{cvprblue}{rgb}{0.21,0.49,0.74}
\usepackage[pagebackref,breaklinks,colorlinks,citecolor=cvprblue]{hyperref}

\def\paperID{4868} 
\def\confName{CVPR}
\def\confYear{2024}

\title{GSNeRF: Generalizable Semantic Neural Radiance Fields\\
with Enhanced 3D Scene Understanding}

\author{Zi-Ting Chou*$^1$\textcolor{white}{\thanks{Equal Contribution}}, \quad Sheng-Yu Huang*$^{1}$, \quad 
I-Jieh Liu$^{1}$, \quad Yu-Chiang Frank Wang$^{1, 2}$, \\
\normalsize \textsuperscript{1} Graduate Institute of Communication Engineering, National Taiwan University \quad \textsuperscript{2} NVIDIA, Taiwan\\
{\tt\small \{r11942101, f08942095, r11942087\}@ntu.edu.tw frankwang@nvidia.com}
}

\begin{document}
\maketitle
\begin{abstract}
Utilizing multi-view inputs to synthesize novel-view images, Neural Radiance Fields (NeRF) have emerged as a popular research topic in 3D vision. In this work, we introduce a \textit{Generalizable Semantic Neural Radiance Fields (GSNeRF)}, which uniquely takes image semantics into the synthesis process so that both novel view image and the associated semantic maps can be produced for unseen scenes. Our GSNeRF is composed of two stages: Semantic Geo-Reasoning and Depth-Guided Visual rendering. The former is able to observe multi-view image inputs to extract semantic and geometry features from a scene. Guided by the resulting image geometry information, the latter performs both image and semantic rendering with improved performances. Our experiments not only confirm that GSNeRF performs favorably against prior works on both novel-view image and semantic segmentation synthesis but the effectiveness of our sampling strategy for visual rendering is further verified.
\end{abstract}    
\section{Introduction}
\label{sec:intro}

3D scene understanding plays a pivotal role in many vision-related tasks, including 3D reconstruction and 3D reasoning. The former domain focuses on low-level understanding, including developing efficient models that interpret and reconstruct observations from various data sources, such as RGB, RGBD, and physical sensors. In contrast, the latter emphasizes high-level understanding, like encoding, segmentation, and common sense knowledge learning. For real-world applications like robotic navigation or augmented reality, both these facets - reconstruction and reasoning - are crucial to enhance interactions with the real world~\cite{garg2020semantics}. Among the challenges in the reconstruction domain, novel view synthesis has consistently been a challenging objective. Being able to generate a novel view image based on a few existing views shows that a model comprehends the scene's geometry and possesses a foundational understanding akin to reconstruction. 

The Neural Radiance Field (NeRF)~\cite{mildenhall2021nerf, barron2021mip, yu2021plenoctrees, muller2022instant, sun2022direct, fridovich2022plenoxels, martin2021nerf, reiser2021kilonerf, chen2022tensorf} has recently risen as an exciting research area, providing a novel way to tackle the task of novel view synthesis. 
By encoding the density and emitted radiance at each spatial location, NeRF is able to compress a scene into a learnable model given several images and the corresponding camera poses of the scene. By incorporating a volumetric rendering skill, images of unseen camera views can be generated with convincing quality. However, NeRF primarily focuses on reconstructing the color information of novel views, and understanding the associated high-level semantic information (\eg, semantic segmentation or object detection) remains a significant challenge.

Recent works~\cite{fu2022panoptic, siddiqui2023panoptic, vora2021nesf, wang2022dm, zhi2021place, kundu2022panoptic} aim to amplify the high-level scene understanding ability of NeRF by integrating NeRF with semantic segmentation. By sharing the information between the semantic object class and their corresponding appearances, these two tasks are capable of benefiting from each other's insights~\cite{liu2023semantic}. 
Nevertheless, these semantic understanding methods stick closely to the original NeRF paradigm, focusing on exploiting the model to represent a specific scene. As mentioned in~\cite{liu2023semantic}, such a strategy requires additional annotations of semantic segmentation maps when applied to a new scene, limiting real-world applicability and generalizability. 

To tackle this problem, generalizable NeRFs~\cite{wang2021ibrnet, yu2021pixelnerf, chen2021mvsnerf, johari2022geonerf, liu2022neural, suhail2022generalizable, wang2022attention} have emerged as a promising solution. They adopt an on-the-fly approach for building a neural radiance field conditioned on extracted features from input images of different scenes rather than encoding the scene representation directly into the model. For instance, PixelNeRF~\cite{yu2021pixelnerf} introduces the idea of conditioning a NeRF model with multi-view images across multiple scenes. This innovation boosts NeRF's ability to generalize to unseen scenes and avoids retraining for each individual scene. Yet, it leaves uncertainties in its application to semantic understanding.

\begin{table*}[t]
\centering
\resizebox{0.8\textwidth}{!}{%
\begin{tabular}{l|cccc}
\toprule
\rule{0pt}{10pt}Method & generalized NeRF & generalized segmentation & sampling strategy & efficient sampling \\ \midrule
Semantic NeRF~\cite{zhi2021place} & - & - & hierarchical & \halfcheckmark \\
Panoptic Neural Fields~\cite{kundu2022panoptic} & - & - & uniform & - \\
Panoptic Lifting~\cite{siddiqui2023panoptic} & - & - & uniform & - \\
DM-NeRF~\cite{wang2022dm} & - & - & hierarchical & \halfcheckmark \\ 
NeSF~\cite{vora2021nesf} & - & \cmark & hierarchical & \halfcheckmark \\ 
Semantic Ray~\cite{liu2023semantic} & \cmark & \cmark & hierarchical & \halfcheckmark  \\ \midrule
\renewcommand{\arraystretch}{1.3}
\rule{0pt}{10pt}Ours & \cmark & \cmark & depth-guided & \cmark \\ \bottomrule
\end{tabular}}
\caption{\label{compareTable}\textbf{Comparisons of semantic-aware NeRFs (i.e., NeRFs, which jointly perform novel view synthesis and semantic segmentation).} Note that NeSF only performs generalizable segmentation, yet its view synthesis remains restricted to the scene-specific training scheme. And, Semantic Ray requires ground truth depth as additional inputs for both training and inference.}
\end{table*}

As a pioneer in integrating a generalizable NeRF with semantic segmentation capabilities, Semantic-Ray (S-Ray)~\cite{liu2023semantic} made notable advancements by introducing a generalizable semantic field concept. By incorporating a cross-reprojection attention module, S-Ray aggregates information of multi-view images from all points sampled along a ray and predicts RGB values and semantic labels. Although this approach appears to capture the necessary contextual information for semantic segmentation, aggregating details of all points introduces noisy features that could interfere with segmentation tasks, as segmentation is mainly concerned with the first object class which the ray encounters. Additionally, their hierarchical sampling approach further additional computation and model complexity in view synthesis. Thus, a properly-designed rendering strategy for a generalizable semantic field would be desirable. 

In this paper, we propose Generalizable Semantic Neural Raidance Fields (GSNeRF) to jointly tackle the problems of generalized novel view synthesis and semantic segmentation. Given a set of source images of a scene with corresponding camera views and a novel target camera view, our GSNeRF is able to derive visual features and depth map predictions of each source view, which are used to predict the depth map of the target view. With the depth map of the target view and properly derived visual features of source views, the novel view RGB image and semantic segmentation can be derived accordingly. Our GSNeRF consists of two key learning stages: \textit{Semantic Geo-Reasoning} and \textit{Depth-Guided Visual Rendering}. The former is to derive visual features and aggregate the predicted depth information of each source view to estimate the depth of the novel view, while the latter renders the RGB image and semantic segmentation map of the target view. We further design two distinct sampling strategies to minimize noisy features and enhance rendering efficiency. Through conducting extensive experiments on both real-world and synthetic datasets, 
we show that our method outperforms current state-of-the-art generalizable NeRF methods in novel view synthesis and semantic segmentation.

The key contributions of our work are as follows:
\begin{itemize}
\item We propose GSNeRF for jointly rendering novel view images and producing the associated semantic segmentation mask on unseen scenes.
\item The proposed Semantic Geo-Reasoning stage learns color, geometry, and semantic information of the input scene, introducing generalization ability of our GSNeRF.
\item Based on the inferred geometry information, the introduced Depth-Guided Visual Rendering stage customizes two different sampling strategies according to the predicted target view depth map, so that image and semantic map rendering can be performed simultaneously.
\end{itemize}
\section{Related Work}
\label{sec:related}

\subsection{Neural Radiance Fields}

The Neural Radiance Field ~\cite{mildenhall2021nerf} has emerged as a widely embraced implicit representation by encoding a 3D scene within a neural network. Nonetheless, as mentioned in~\cite{deng2022depth}, the original NeRF method demands extensive training time, ranging from hours to days, and relies on dozens of multi-view images as input. As a result, a large number of subsequent works~\cite{muller2022instant, sun2022direct, yu2021plenoctrees, fridovich2022plenoxels} are put forward to confront these issues. To shorten the training process, methods such as Instant NGP~\cite{muller2022instant} and DVGO~\cite{sun2022direct} opt for a balance between speed and memory, using hash encoding and voxel encoding to minimize the training time to minutes. On the other hand, to address the demand for multiple input views, some methods~\cite{roessle2022dense, johari2022geonerf, deng2022depth} introduce depth as additional supervision, a notable example being DS-NeRF~\cite{deng2022depth}. DS-NeRF applies extra depth supervision to ensure that the NeRF accurately encodes a scene's geometry, thus enabling the construction of neural radiance fields with fewer images. This method shows how incorporating depth information can enhance the quality of novel view synthesis while reducing the number of required source images.

\subsection{Generalizable Novel View Synthesis}

Since NeRF directly encodes information of a scene into a neural network, it is only applicable to the training scene of interest. In response to this limitation, some methods~\cite{trevithick2021grf, yu2021pixelnerf, chen2021mvsnerf, johari2022geonerf, liu2022neural, suhail2022generalizable, wang2022attention} propose on-the-fly construction of NeRF, allowing trained NeRFs applied for synthesizing novel views in unseen scenes. The concept is to extract features from each source view image and condition the NeRF renderer on these scene-specific features, typically through~\cite{yu2021pixelnerf} 2D CNN or~\cite{chen2021mvsnerf, johari2022geonerf} cost volume techniques that aggregate features across different images. While these methods realize generalizable novel view synthesis, extensions of such NeRFs to high-level scene understanding tasks remainsan open challenge.
\begin{figure*}[t]
	\centering
	\includegraphics[width=1.85\columnwidth]{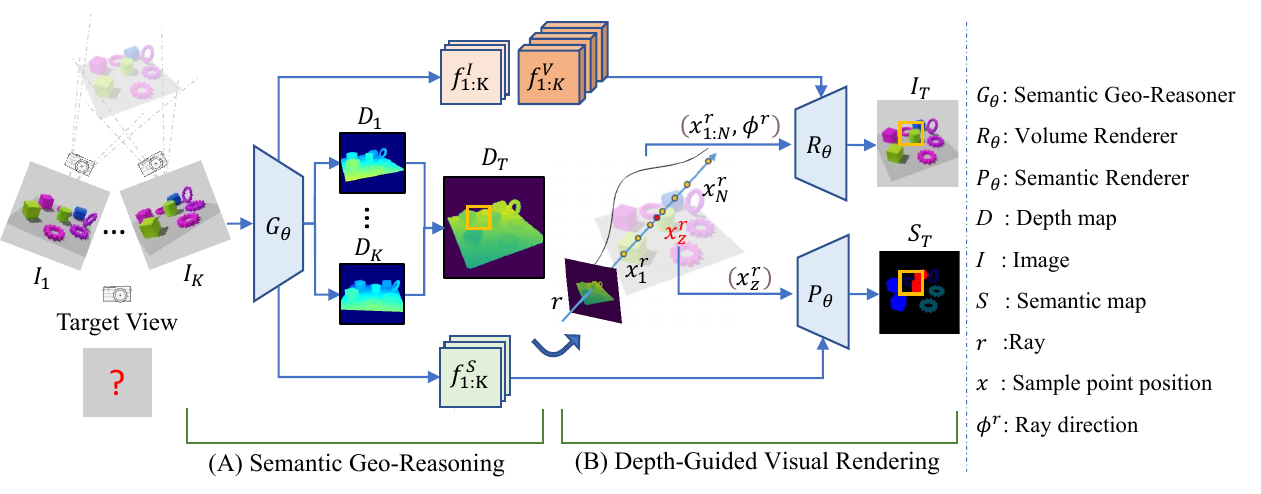}
    \caption{\textbf{Overview of GSNeRF}: including \textit{Semantic Geo-Reasoning} and \textit{Depth-Guided Visual Rendering}. Given K multi-view image $I_{1:K}$ of a scene, the Semantic Geo-Reasoner predicts the depth map $D_{1:K}$ for each source image, which is aggregated to estimate the target view depth map $D_T$. With $D_T$ as key geometric guidance, we design Depth-Guided Visual Rendering to render target view image $I_T$ and semantic segmentation $S_T$, by Volume Renderer $R_\theta$ and Semantic Renderer $P_\theta$, respectively. 
    }
    \vspace{-3mm}
	\label{fig:archi}
\end{figure*}


\subsection{Multi-tasking NeRF}

Recently, several methods~\cite{zhi2021place,vora2021nesf,kundu2022panoptic,fu2022panoptic,wang2022dm,siddiqui2023panoptic,liu2023semantic} seek to enhance NeRF with higher-level understanding abilities. Semantic NeRF~\cite{zhi2021place} stands as the first method to integrate semantic segmentation with NeRF by designing two projection heads for predicting both semantics and color simultaneously. Although it achieves satisfactory performance and robustness on semantic segmentation tasks, it requires retraining for each scene. This implies a need for not only RGB images but also corresponding semantic segmentation maps to represent each new scene, hindering their generalizability. NeSF~\cite{vora2021nesf} has made an attempt to improve this aspect by proposing a half-generalizable method that includes a generalizable semantic segmentation module. That is, after training an original NeRF of each scene, they extract a 3D voxel-like grid density from the neural radiance field and train a generalizable 3D UNet to do a semantic segmentation task. However, this learning scheme still requires additional time to retrain NeRFs for novel scenes. 

S-Ray~\cite{liu2023semantic} advances further by proposing a generalized semantic neural field. Following the backbone of Neuray~\cite{liu2022neural}, they propose a cross-reprojection attention module. Given a ray emitted from the target viewpoint, they uniformly sample points along the ray, and hierarchically sample more points on the second forward pass according to the density predicted from the first pass. With the cross-reprojection module, they aggregate the contextual information on all sampled points. Despite achieving good results on both real-world and synthetic datasets, the sampling strategy is time-consuming. Also, aggregating all information along the entire ray can introduce noisy features that may hamper segmentation tasks, as segmentation primarily focuses on the first object class encountered by the ray.

Contrarily, our proposed  GSNeRF aims to predict the depth map of the target view and utilize it to sample points along the ray more efficiently. This unique strategy enables us to sample points focusing on the first encountered object and filter out unrelated noisy points for semantic segmentation prediction. In Table~\ref{compareTable}, we compare the characteristics of recent semantic-aware NeRFs with GSNeRF.

\begin{figure}[t]
	\centering
	\includegraphics[width=1\columnwidth]{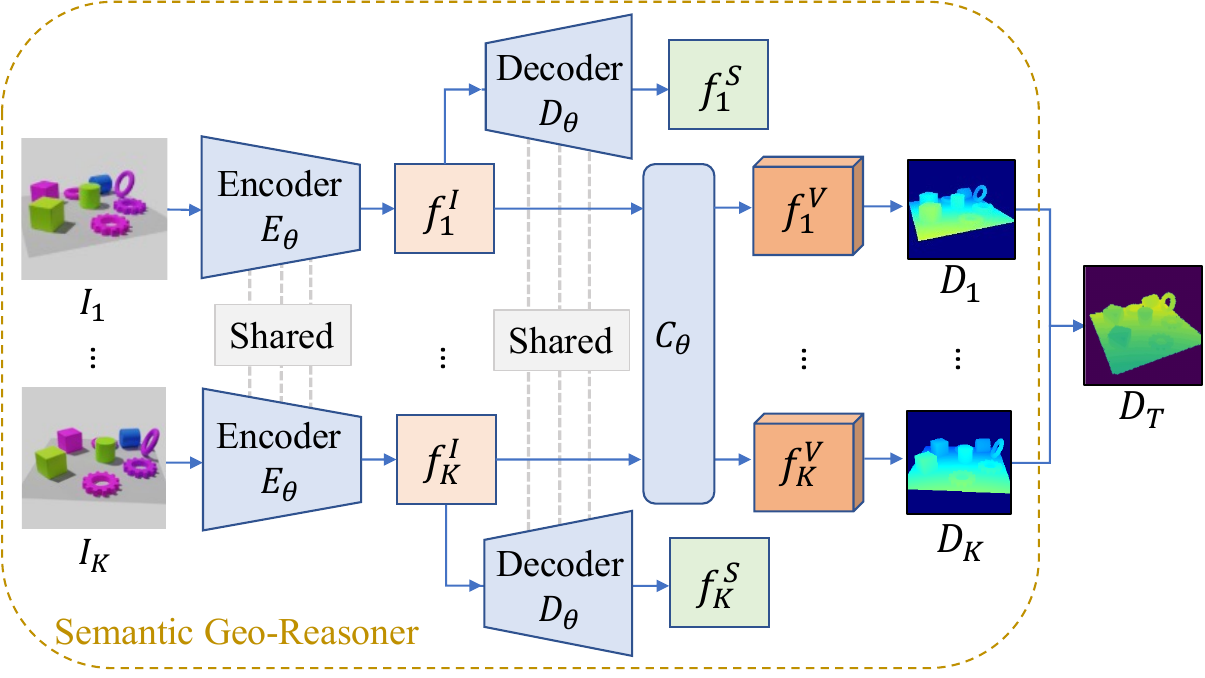}
    \caption{\textbf{Semantic Geo-Reasoning.}
    The Semantic Geo-Reasoner contains a shared Encoder $E_{\theta}$, a shared Decoder $D_{\theta}$, and a cost-volume aggregator $C_{\theta}$, producing image and volume features ($f^I_{1:K}$ and $f^V_{1:K}$), with the associated depth maps $D_{1:K}$. The depth map $D_T$ of the target view are estimated from $D_{1:K}$.
    }
    \vspace{-3mm}
	\label{fig:cost_archi}
\end{figure}
\section{Brief Review of Generalizable NeRFs}
\label{sec:review}
For the sake of completeness, we provide a brief review of generalizable NeRFs~\cite{trevithick2021grf, yu2021pixelnerf, chen2021mvsnerf, johari2022geonerf, liu2022neural}. Given $K$ multi-view images $I_{1:K} = \{I_1, I_2, ..., I_K\}$ with corresponding camera poses $\xi_{1:K} = \{\xi_1, \xi_2, ..., \xi_K\}$ of a scene, and a target camera pose $\xi_T$, the goal is to synthesize the novel view image $I_T$. Rays are sampled across the scene from the target camera pose to render the novel view image. Along each ray, $r$, $N_c$ points (coarse points) are sampled within the given near bound $b_l$ and far bound $b_u$. Existing methods partition $[b_l, b_u]$ into $N_c$ bins uniformly and draw one sample within each bin as defined by:
\begin{equation} \label{eq:uniform}
    t_n^r \sim \mathcal{U} [b_l + \dfrac{n-1}{N_c}(b_u-b_l), b_l + \dfrac{n}{N_c}(b_u-b_l)],
\end{equation}
where $t^r_n$ is the $n$-th sample along $r$.
With the corresponding camera position  $o^r$ and viewing directions $\phi^r$, the $n$-th sampled point $x_n^r$ can be derived as $x_n^r = o^r + t_n^r\phi^r$, with $x_n^r \in \mathbb{R}^3$. Several methods~\cite{trevithick2021grf, yu2021pixelnerf, chen2021mvsnerf, liu2022neural} also use hierarchical sampling, conducting a second forward pass to sample $N_f$ points (fine points) where the first $N_c$ uniformly sampled points show higher predicted density.

With the above information, a generalized NeRF can be represented by a learnable continuous function $A_\theta$, where $\theta$ denotes the learnable parameters.  $A_\theta$ computes a volumetric density $\mathbf{\sigma}_n^r\in \mathbb{R}^+$ and a view-dependent color representation $\mathbf{c}_n^r\in \mathbb{R}^3$ for each point $x^r_n$. That is,
\begin{equation}\label{eq:obj_func}
    \mathbf{\sigma}_n^r, \mathbf{c}_n^r = A_\theta(x_n^r, \phi^r, f_{1:K}(w(x_n^r))),
\end{equation}
where $f_{1:K}$ are features of $I_{1:K}$ and $w(.)$ is the function that reprojects the 3D position $x_n^r$ onto the image plane with known camera poses.

To this end, the color value $\mathbf{C}(r)$ of the pixel corresponding to the ray $r$ is formulated as:
\begin{equation}
\label{eq:color render}
\mathbf{C}(r)=\sum_{n=1}^{N_c} T_n^r\, \alpha_n^r\, \mathbf{c}_n^r.
\end{equation}
\begin{equation}
T_n^r=exp(-\sum_{i=1}^{n-1} \sigma_i^r\delta_i^r), \;\;\;\;\;\;\;
\alpha_n^r = 1-exp (-\sigma_n^r\delta_n^r),
\end{equation}
where $T_n^r$ is the accumulated transmittance and $\delta_n^r = ||x_{n+1}^r - x_{n}^r||_2$ represents the distance between every two adjacent points along $r$.

As is the case with other NeRF methods, a generalized NeRF also optimizes the model based on the rendering loss between the rendered image and the GT image:
\begin{equation} \label{Limage}
    \mathcal{L}_{image} = \sum_{r\in R}\| \mathbf{C}(r) - \mathbf{\hat{C}}(r)\|_2^2,
\end{equation}
where $\hat{\mathbf{C}}(r)$ represents the GT color of the renderer image corresponding to a ray $r$.

Our method is built on top of generalized NeRF methods but with a more efficient sampling strategy that can deal with both semantic segmentation and image rendering.
\section{Method}
\subsection{Problem Formulation and Model Overview}
We first define the setting and notations used in this paper. Given a scene with $K$ multi-view images $I_{1:K}$ and their corresponding camera poses $\xi_{1:K}$ along with a target camera pose as input, we aim to achieve novel-view synthesis and semantic segmentation simultaneously. During training, we synthesize both the novel view image $I_T$ and the semantic segmentation map $S_T$ from $\xi_T$, given ground truth (GT) image $\hat{I}_T$ and semantic segmentation map $\hat{S}_T$ of the target view $\xi_T$ Once the model completes training on a given set of training scenes, it is able to directly generalize to unseen scenes without finetuning. 

To achieve this goal, we propose Generalizable Semantic Neural Radiance Fields (GSNeRF), as illustrated in Figure~\ref{fig:archi}. GSNeRF consists of two key learning stages: Semantic Geo-Reasoning and Depth-Guided Visual Rendering. For Semantic Geo-Reasoning, we employ a Semantic Geo-Reasoner $G_\theta$ on each input source image $I_k \in \mathbb{R}^{H\times W\times 3}$ to create 2D feature $f^I_k \in \mathbb{R}^{H\times W\times d}$, semantic feature $f^S_k \in \mathbb{R}^{H\times W\times d}$, 3D volume feature $f^V_k \in \mathbb{R}^{H\times W\times L\times d}$, and depth predictions $D_k \in \mathbb{R}^{H\times W}$, where $H$ and $W$ represent the height and width of the image, $L$ is the depth sample number defined in~\cite{yao2018mvsnet} to determine the volume size for the 3D feature, and $d$ is the feature dimension. The source view depth maps are integrated to estimate the depth for the target view $D_T$. As for Depth-Guided Visual Rendering, we conduct a unique sampling strategy based on $D_T$ to minimize noisy features and enhance rendering efficiency. The sampled points, along with the previously extracted feature, are respectively input into the volume renderer $R_\theta$ and semantic renderer $P_\theta$, synthesizing the target view's image and semantic segmentation map. In the following subsection, we will explain the Semantic Geo-Reasoning and Depth-guided Visual Rendering process within our GSNeRF framework. 

\subsection{Generalizable Semantic NeRF}
\subsubsection{Semantic Geo-Reasoning}\label{sec:semgeoreasoner}
Given $K$ multi-view source images $I_{1:K}$ of a scene, our Semantic Geo-Reasoner $G_{\theta}$ in Figure~\ref{fig:archi} is designed to achieve two goals. First, geometric clues are extracted from $G_{\theta}$, comprising 3D Volume features \(f^V_{1:K}\) and depth maps $D_{1:K}$ for each source image. In addition, 2D image features $f^I_{1:K}$ and semantic features $f^S_{1:K}$ for each source image can be jointly produced. Secondly, our $G_{\theta}$ is learned to predict the target view depth map $D_T$, which is realized by deterministic aggregation and reprojection of all source view depth predictions $D_{1:K}$. We now detail this learning stage.

As depicted in Figure~\ref{fig:cost_archi}, our Semantic Geo-Reasoner consists of a shared 2D CNN encoder $E_\theta$ that extracts $f^I_{1:K}$ from every image $I_{1:K}$. To construct geometry-aware features, a cost volume aggregator $C_\theta$~\cite{gu2020cascade, yao2018mvsnet, huang2021m3vsnet, dai2019mvs2} is applied to gather 2D features $f^I_{1:K}$ across different images. As a result, $C_\theta$ produces 3D feature $f^V_{1:K}$, which are ultimately used to predict the depth map $D_{1:K}$ for every source view image. To obtain semantic-aware features for semantic prediction, a shared semantic feature decoder $D_\theta$ is additionally applied to derive 2D semantic features $f^S_{1:K}$ from $f^I_{1:K}$. 

Since $D_{1:K}$ is a vital intermediate element that assists in estimating $D_T$, which allows the subsequent depth-guided rendering process, we strive to ensure the precise prediction of $D_{1:K}$, whether or not the ground truth depth images are accessible. To achieve this, we integrate ideas from DSNeRF~\cite{deng2022depth} and GeoNeRF~\cite{johari2022geonerf} to supervise our $G_{\theta}$ with either a depth-supervised loss or a self-supervised depth loss. If the ground-truth depth maps of $D_{1:K}$ are available, our depth-supervised loss $\mathcal{L}_D$ is given as:

\begin{equation}
    \mathcal{L}_D = \dfrac{1}{K}(\sum_{k=1}^K\| D_k - \hat{D}_k\|_{s1}),
\end{equation}
Here, $\hat{D}_k$ represents the GT depth map of view $k$, $\|.\|_{s1}$ represents smooth L1 loss~\cite{girshick2015fast}. If $\hat{D}_k$ is unavailable, we use a self-supervised depth loss $\mathcal{L}_{ssl}$ to regularize our depth predictions by considering cross-view depth consistency between all source views. Following the self-supervised depth estimation approach of RCMVSNet~\cite{chang2022rc}, $\mathcal{L}_{ssl}$ is defined as:
\begin{equation} \label{eq:ssloss}
    \mathcal{L}_{ssl}  = \lambda_1\mathcal{L}_{RC} + \lambda_2\mathcal{L}_{SSIM} + \lambda_3\mathcal{L}_{Smooth},  
\end{equation}
where $\mathcal{L}_{RC}$, $\mathcal{L}_{SSIM}$, and $\mathcal{L}_{Smooth}$ represent reconstruction, structure similarity, and depth smoothness losses~\cite{chang2022rc}. The hyper-parameters $\lambda_1$, $\lambda_2$, and $\lambda_3$ are also selected following~\cite{chang2022rc} (see supplementary material for details).

To this end, the target view depth map $D_T$ is calculated using \(D_{1:K}\) through a deterministic algorithm. In simple terms, we directly project the pixels of each depth map into 3D space and then reproject them back to the target view. More details about the self-supervised depth loss and corresponding pseudo-code of target view depth estimation are available in the supplementary material.
\renewcommand{\arraystretch}{1.3}
\begin{table*}[t] 
\centering
\resizebox{\textwidth}{!}{%
\begin{tabular}{l|cc|ccccc|ccccc}
\toprule
\multirow{2}{*}{Generalized method} & \multicolumn{2}{c|}{GT Depth} & \multicolumn{5}{c|}{ScanNet~\cite{dai2017scannet}}   & \multicolumn{5}{c}{Replica~\cite{straub2019replica}}                 
\\ \cmidrule{2-13} 
      & \multicolumn{2}{c|}{Train / Test} & mIoU  & acc. / class acc. & PSNR$\uparrow$  & SSIM$\uparrow$  & LPIPS$\downarrow$ & mIoU  & acc. / class acc. & PSNR$\uparrow$  & SSIM$\uparrow$  & LPIPS$\downarrow$ 
      \\ \midrule
Neuray~\cite{liu2022neural} + semhead      & \multicolumn{2}{c|}{\cmark\ /\ \cmark}       & 52.09 & 67.81 / 61.98 & 25.01 & 83.07 & 31.63 & 44.37 & 79.93 / 54.25 & 26.21 & 87.37 & 30.93 \\
GeoNeRF~\cite{johari2022geonerf} + semhead   & \multicolumn{2}{c|}{\cmark\ /\ \,\,\,\,\, }  & 53.78 & 76.18 / 61.90 & \textbf{32.55} & \textbf{90.88} & 12.69 & 45.12 & 81.67 / 52.36 & 28.70 & 88.94 & 20.42 \\
S-Ray~\cite{liu2023semantic} & \multicolumn{2}{c|}{\cmark\ /\ \cmark}     & 55.53 & 77.79 / 60.92   & 25.19 & 83.66 & 30.98 & 45.30  & 80.48 / 53.72   & 26.38 & 88.13 & 30.04 \\
\rowcolor{black!20}GSNeRF (Ours)  & \multicolumn{2}{c|}{\cmark\ /\ \,\,\,\,\, }  & \textbf{58.30}  & \textbf{79.79 / 65.93}   & 31.33 & 90.73 & \textbf{12.53} & \textbf{51.52} & \textbf{83.41 / 61.29}   & \textbf{31.16} &  \textbf{92.44} & \textbf{12.54} 
\\ \midrule
MVSNeRF~\cite{chen2021mvsnerf} + semhead        &          &         & 43.06 & 66.90 / 53.63  & 24.14 & 80.36 & 34.63 & 30.21 & 69.35 / 39.75 & 23.68 & 84.37 & 28.08 \\
GeoNeRF~\cite{johari2022geonerf} + semhead        &          &         & 45.11 & 67.12 / 53.44 & 30.75 & 88.27 & 16.48 & 40.35 & 74.63 / 49.18 & 29.92 & 91.14 & 17.60 \\
GNT~\cite{wang2022attention} + semhead            &          &         & 43.49 & 62.06 / 51.84 & 24.39 & 82.37 & 28.36 & 38.14 & 71.44 / 47.46 & 24.56 & 87.31 & 20.97 \\
Neuray~\cite{liu2022neural} + semhead         &          &         & 46.09 & 66.39 / 53.79 & 25.24 & 84.39 & 31.33 & 40.91 & 76.23 / 50.15 & 27.80  & 89.55 & 23.68 \\
S-Ray~\cite{liu2023semantic}                    &       &       & 47.69 & 64.90 / 54.47    & 25.13 & 84.18 & 30.44 & 43.27 & 77.63 / 52.85   & 26.77 & 88.54 & 22.81 \\
\rowcolor{black!20}GSNeRF (Ours)  &  \multicolumn{2}{c|}{\ \,\,\,\,\, }   & \textbf{52.21} & \textbf{74.71 / 60.14} & \textbf{31.49} & \textbf{90.39} & \textbf{13.87} & \textbf{51.23} & \textbf{83.06 / 61.10}    & \textbf{31.71} & \textbf{92.89} & \textbf{12.93} \\ 
\bottomrule
\end{tabular}%
}
\caption{\label{Table1}\textbf{Quantitative results on ScanNet \& Replica.} Note that methods in the first four rows take GT depth as inputs or training supervision, while the methods in the last six rows do not observe GT depth during training/testing.}
\label{tab:main}
\end{table*}

\subsubsection{Depth-Guided Visual Rendering}

We now discuss how we perform volume and semantic rendering with the guidance of the predicted $D_T$. That is, given a target view $\xi_T$, a ray is sampled for each pixel on the image plane to render the corresponding RGB value for $I_T$ and the semantic segmentation output for $S_T$ via the Volume Renderer $R_{\theta}$ and the Semantic Renderer $P_{\theta}$.

\paragraph{Volume Rendering}
For \textbf{\textit{volume rendering of $I_T$}}, existing semantic-NeRF approaches~\cite{zhi2021place, vora2021nesf, liu2023semantic} follow conventional NeRFs~\cite{mildenhall2021nerf} by applying either uniform sampling or hierarchical sampling along each ray $r$, as depicted in Table~\ref{compareTable}. However, both strategies require dense sampling to locate the surface of objects in the given scene, limiting the sampling efficiency. To promote the efficiency of our sampling process, we conduct a depth-guided sampling strategy by focusing on sampling points near the depth values of $D_T$. More specifically, this is achieved by modifying the sampling strategy in Eq.~\ref{eq:uniform} as follows:
\begin{align}
t_n^r &\sim \mathcal{G}(z, v^2), & v = &\dfrac{\text{min}(\|z-b_u\|,\|z-b_l\|)}{3},\label{eq:gaussian}
\end{align}
where $t_n^r$ is the $n$-th sample along the ray $r$, and $\mathcal{G}$ represents Gaussian (normal) distribution. Note that $z$ and $v$ represent the mean and standard deviation of the distribution, respectively, while the value of $z$ is determined as the predicted depth for the pixel $p$ that emits the ray $r$  (\ie, $z = D_T(p)$). By these sampling point values, we derive $N$ sampled points $x^r_{1:N}$ (\ie,  $x_n^r = o^r + t_n^r\phi^r$) closer to the surface of the first encountered object by $r$, avoiding considering noisy points that are far from the object surface. Please see the supplementary material for further details.

We feed all these points alongside the 2D feature $f^I_{1:K}$ and 3D volume feature $f^V_{1:K}$ as additional conditions into the volume renderer $R_\theta$ to predict the novel view image of the scene. Specifically, we reproject the point $x^r_{n}$ onto each source view to get aggregated features from these condition features. Thus, we write the aggregated feature $f_{n,k} \in \mathbb{R}^{2d}$ of the $n$-th sampled point $x^r_{n}$ from the $k$-th view as:
\begin{equation}
    f_{n,k} = [f^I_k(w(x^r_n)); f^V_k(x^r_n)], \forall{k \in [1:K]}
\end{equation}
where $w(\cdot)$ is the reprojection function used in Eq.~\ref{eq:obj_func} and $[\cdot\,;\,\cdot]$ denotes concatenation along feature dimension. We further define the global feature $f_{n,0}$ as guidance across all $K$ source views as:
\begin{equation}
    f_{n,0} = [mean(\{f_{n,k}\}_{k=1}^K);var(\{f_{n,k}\}_{k=1}^K)].
    \label{eq:mean}
\end{equation}
Following GeoNeRF~\cite{johari2022geonerf} methodology, the volume renderer $R_\theta$ (with learnable parameters $\theta$) predicts per point density and radiance based on the feature $f_n = \{f_{n,k}\}_{k=0}^K\ \in \mathbb{R}^{(K+1)\times 2d}$ mentioned above:
\begin{equation} \label{eq:volume_renderer}
\mathbf{\sigma}_n^r, \mathbf{c}_n^r = R_\theta(x_n^r, \phi^r, f_n, M_n),
\end{equation}
where $M_n = \{M_{n,k}\}_{k=0}^K$ denotes the mask used to mask out unrelated features of a view. If the sampled point $x_n^r$ lies behind the predicted depth map $D_k$ or if the reprojected location is outside the source image,  $M_{n,k}$ is set to 0. However, the mask for the global feature $M_{n, 0}$ is always set to~1.

To this end, we apply the same volume rendering as in Eq.~\ref{eq:color render}, predicting the target view Image $I_T$. We also apply $\mathcal{L}_{image}$ as in Eq.~\ref{Limage} to supervise the rendered image quality.

\paragraph{Semantic Rendering}
\textbf{\textit{Semantic rendering of $S_T$}}, on the other hand, is different from the above volume rendering process. Specifically, image rendering considers the color of both the surface and nearby objects because of possible semi-transparent or transparent objects. Conversely, semantic segmentation focuses only on the first object a ray touches. Therefore, for each ray $r$, we sample only \textit{\textbf{one}} surface point $x_z^r$ estimated by the target depth map for each ray $r$ to render its semantic value (\ie, $x^r_z = o^r + z\phi^r$, where $z$ is the same depth value as mentioned in Eq.~\ref{eq:gaussian}). More specifically, with the aggregated semantic feature $f_z = \{f^S_k(w(x^r_z))\}_{k=1}^K \in \mathbb{R}^{K\times d}$, our semantic renderer $P_\theta$  with learnable parameters $\theta$ predicts the semantic logits $\mathbf{S}(r)$ of ray $r$ as follows:
\begin{equation} \label{eq:sem_render}
    \mathbf{S}(r) = P_\theta(x^r_z, f'_{z}, M),
\end{equation}
where
\begin{equation}
    f'_{z} = f_z + \{f_0\}\in \mathbb{R}^{(K+1)\times d}. \\ 
\end{equation}
Note that we follow volume rendering in Eq.~\ref{eq:mean} to define a global semantic feature $f_0$ as guidance across all views:
\begin{equation}
    f_0 = [mean(f_z);var(f_z)]. 
\end{equation}
As for the mask $M = \{M_k\}_{k=0}^K$ in Eqn.~\ref{eq:sem_render}, it is similar to the one used in Eq.~\ref{eq:volume_renderer}, where $M_{k}$ is to mask out the unrelated features of a view if the sample point $x^r_z$ is behind the corresponding depth value of $D_k$.

Eventually, we utilize a semantic loss to supervise the novel view semantic prediction. Here, we use a multi-class cross-entropy loss to reduce the semantic prediction error:
\begin{equation}
    \mathcal{L}_{sem} = \sum_{r\in R}(\mathbf{S}(r)\text{log}\mathbf{\hat{S}}(r)),
\end{equation}
where $\hat{S}(r)$ represent the ground truth segmentation score along the ray $r$.

\begin{figure*}[t]
\Huge
\resizebox{\textwidth}{!}{%

\begin{tabular}{cc|c|cc|c}
 {\fontsize{40}{48}\selectfont S-Ray} & 
 {\fontsize{40}{48}\selectfont GSNeRF (Ours)} &
 {\fontsize{40}{48}\selectfont GT} &
 {\fontsize{40}{48}\selectfont S-Ray} &
 {\fontsize{40}{48}\selectfont GSNeRF (Ours)} & 
 {\fontsize{40}{48}\selectfont GT} \\
 
\includegraphics[]{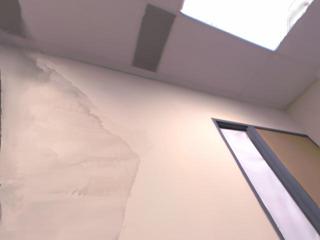} &
\includegraphics[]{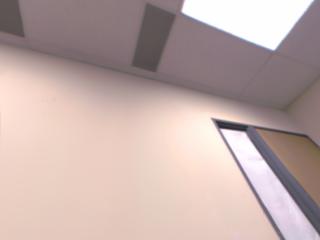} &
\includegraphics[]{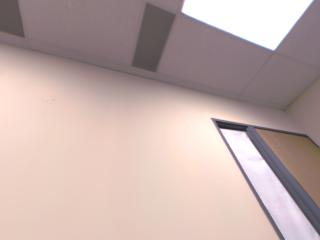} &
\includegraphics[]{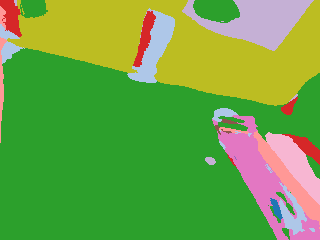} &
\includegraphics[]{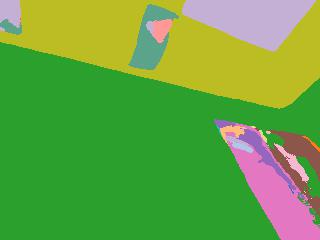} & 
\includegraphics[]{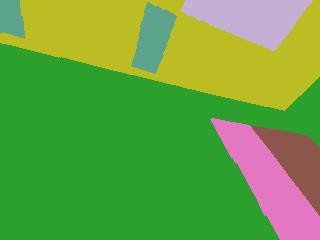} \\

\includegraphics[]{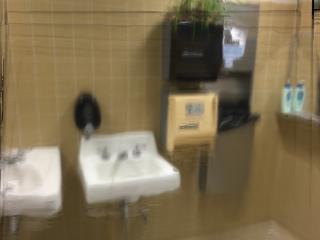} &
\includegraphics[]{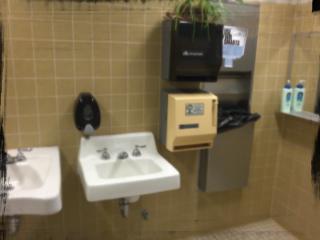} &
\includegraphics[]{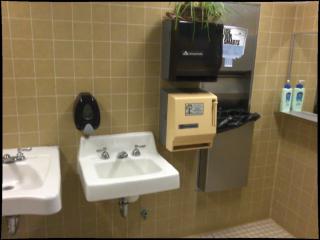} &
\includegraphics[]{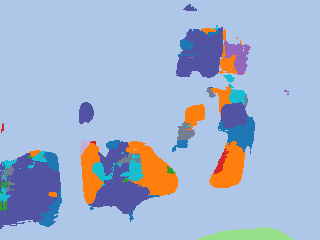} &
\includegraphics[]{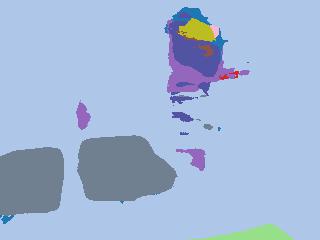} &
\includegraphics[]{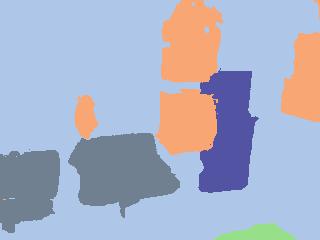}  \\

\includegraphics[]{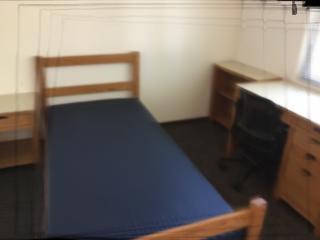} &
\includegraphics[]{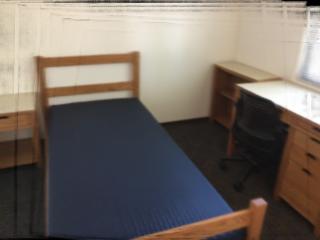} &
\includegraphics[]{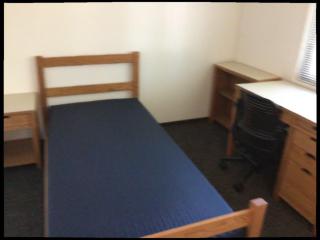} &
\includegraphics[]{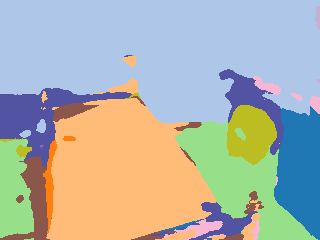} &
\includegraphics[]{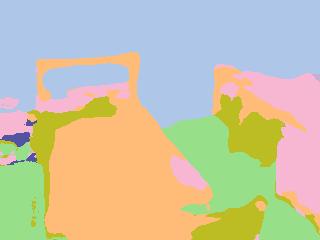} &
\includegraphics[]{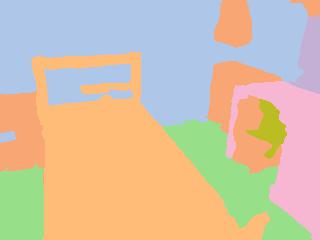}  \\

\includegraphics[]{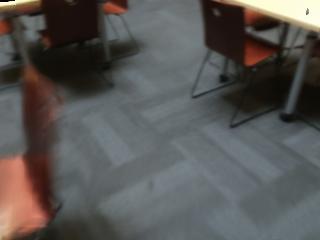} &
\includegraphics[]{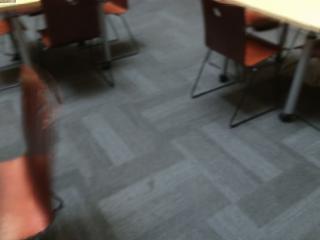} &
\includegraphics[]{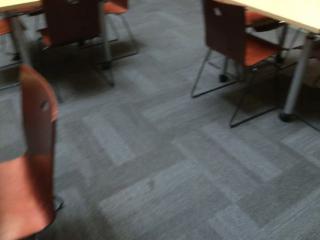} &
\includegraphics[]{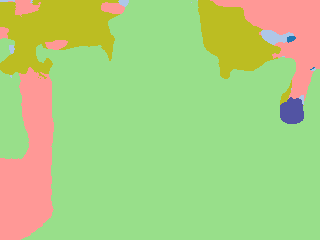} &
\includegraphics[]{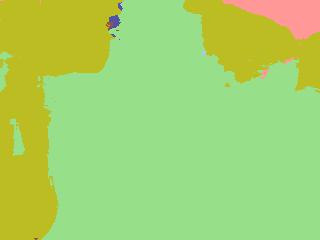} &
\includegraphics[]{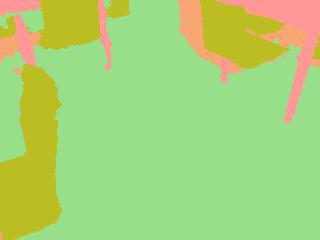}  \\

\end{tabular}}

\caption{\textbf{Qualitative evaluation.} We compare the visual quality of the rendered novel view images (the first three columns) and semantic segmentation maps (the last three columns) with S-Ray~\cite{liu2023semantic}.}
\label{fig:quality}
\vspace{-2mm}
\end{figure*}

\subsection{Training and Inference}
\subsubsection{Training}
At the training stage, the total loss is calculated by combining the losses that are mentioned above. The aggregated loss function for our model, when ground-truth depth is available, is given by:
\begin{equation}
\mathcal{L} = \mathcal{L}_{image} + \mathcal{L}_D + \lambda\mathcal{L}_{sem}.
\end{equation}
If the GT depth is \textit{not} available, our model is trained by:
\begin{equation} \label{eq:no_dpeth_loss}
\mathcal{L} = \mathcal{L}_{image} + \mathcal{L}_{ssl} + \lambda\mathcal{L}_{sem}  ,
\end{equation}
where the coefficient $\lambda$ is set to 0.5 for simplicity.

In essence, this optimization process enables our model to estimate the depth map of the target camera view accurately. Utilizing this depth information, the model is capable of efficiently sampling points to render new view images and create corresponding semantic segmentation maps.

\subsubsection{Inference}
At the inference stage, our GSNeRF is able to generalize to unseen scenes without the need for retraining from scratch. Due to the design of the conditioning renderer on scene-specific features $f^I$, $f^V$, and $f^S$, our model is able to build a semantic neural radiance field on the fly. Once we finish training our model on training scenes, we are able to directly infer novel view images and semantic segmentation maps on any scene given multi-view images and corresponding camera poses of that scene.

\section{Experiments}

\subsection{Datasets}
To evaluate the effectiveness of our proposed method, we conduct experiments on both real-world and synthetic datasets. For real-world data, we use ScanNet~\cite{dai2017scannet}, a large-scale, indoor RGB-D video dataset with over 2.5 million views from 1513 distinct scenes, including semantic annotations and camera poses. Following the setting presented in S-Ray~\cite{liu2023semantic}, we train our model on 60 scenes and test its generalization on 10 new, unseen scenes. For synthetic data, we utilize Replica~\cite{straub2019replica}, a 3D reconstruction-based indoor dataset comprising 18 high-quality scenes with dense geometry, HDR textures, and semantic labels. Following the Semantic NeRF~\cite{zhi2021place} supplied rendered data, we train our model on 6 distinct scenes across 12 video sequences and test on 2 novel scenes across 4 video sequences. Please see the supplementary material for details.

\subsection{Results and analysis}

\subsubsection{Quantitative Results}
Table~\ref{tab:main} compares our method against several baselines, in which S-Ray~\cite{liu2023semantic} serves as a primary baseline since it is the first to address semantic segmentation with NeRF in a generalized setting. We further conduct experiments on MVSNeRF~\cite{chen2021mvsnerf}, GeoNeRF~\cite{johari2022geonerf}, GNT~\cite{wang2022attention} and NeuRay~\cite{liu2022neural} with the semantic head as described in~\cite{liu2023semantic}. It is worth noting that, even though S-Ray does not explicitly mention using a depth map of each source view image as input in their paper, their official implementation takes the depth maps $D_{1:K}$ as additional input along with $I_{1:K}$ during training and testing. Therefore, we follow their official implementation to reproduce S-Ray results. 

From Table~\ref{tab:main}, it can be seen that our model generalizes well to unseen scenes. To show that our GSNeRF is also capable of producing impressive results without depth supervision, we conduct an additional experiment in which we only use the loss function in Eq.~\ref{eq:no_dpeth_loss} to train our GSNeRF in Table~\ref{tab:main}. We also conduct experiments of S-Ray, Neuray with only RGB images as input, and GeoNeRF without GT depth as supervision for a fair comparison. From Table~\ref{tab:main}, we see that even without access to GT depth and with the target depth information regularized in a self-supervised manner, our method still outperforms all the other baseline approaches. This suggests that our model can be trained without requiring depth map supervision, confirming the effectiveness and practicality of our proposed approach. We also note that, although GeoNeRF + semhead (with depth supervision) slightly outperforms our approach regarding PSNR and SSIM for the rendered RGB images on ScanNet, our method excels in all semantic segmentation metrics by approximately 5\%. More comprehensive discussions are presented in the supplementary materials.
\begin{table}[t]

  
  \centering
  \resizebox{\linewidth}{!}{%
  \begin{tabular}{l|c|cc|c|c}
\toprule
 & $I_T$ & \multicolumn{2}{c|}{$S_T$} &  &  \\ \midrule
Model & Depth-guided & renderer $P_\theta$ & Depth-guided & PSNR & mIoU \\ \midrule
A & - & - & - & 28.48 & 37.88 \\
B & \cmark & - & - & 30.56 & 41.72 \\
C & \cmark & \cmark & - & 30.88 & 49.47 \\ \midrule
Ours & \cmark & \cmark & \cmark & \textbf{31.49} & \textbf{52.21} \\ \bottomrule
\end{tabular}}
    \caption{\textbf{Ablation studies on GSNeRF.} We verify the effectiveness of using depth-guided rendering for rendering $I_T$ and $S_T$. Note that, without $P_\theta$, we simply take a pre-trained 2D segmentor on $I_T$ to produce $S_T$.}
    \label{tab:ablation_board}
\end{table}

\vspace{-0mm}
\begin{figure}[t]
	\centering
	\includegraphics[width=1\linewidth]{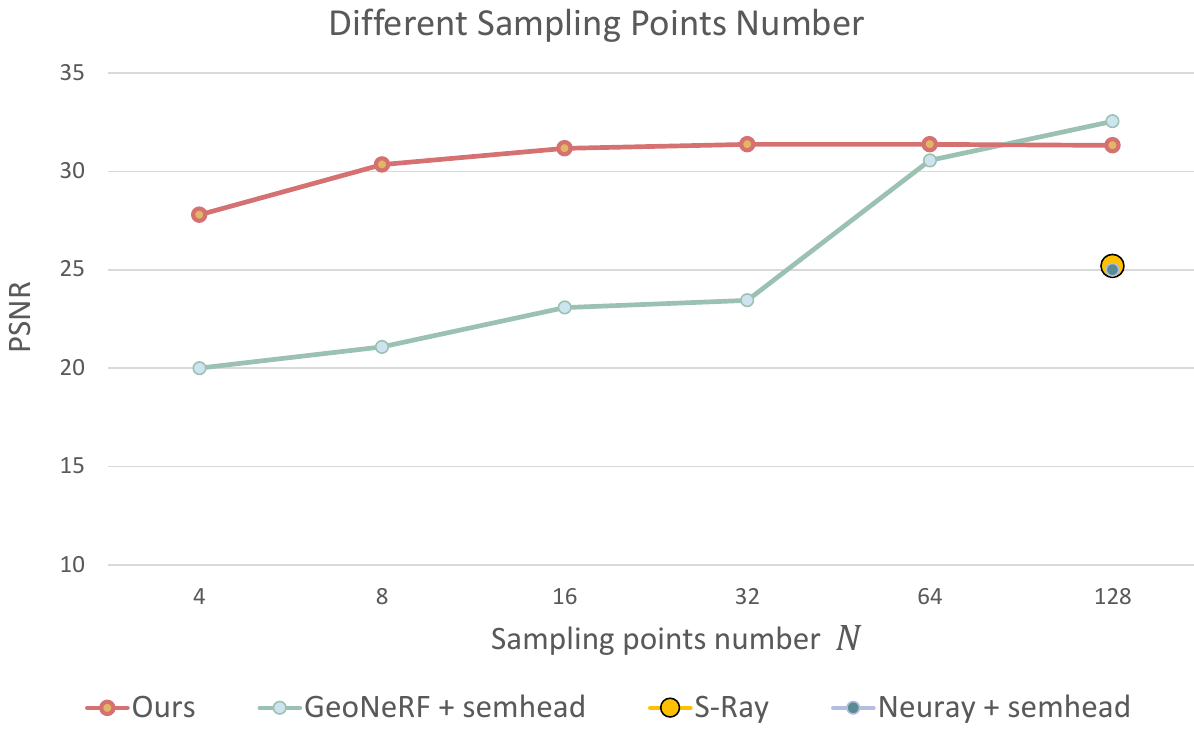}
    \caption{\textbf{Sampling efficiency on ScanNet.} Compared to SOTAs, our GSNeRF is able to achieve significantly improved rendering performance especially when the number of sampling points is small. Even with increasing numbers of sampling points, GSNeRF still performs favorably against existing models.
    }
    \vspace{-3mm}
	\label{fig:points}
\end{figure}

\subsubsection{Qualitative Results}
The qualitative results of our approach compared to S-Ray~\cite{liu2023semantic} are depicted in Fig.~\ref{fig:quality}. This comparison illustrates that while the existing method can approximate the contours of objects, it is unable to capture the geometrical details, causing segmentation inconsistencies. In contrast, our GSNeRF accurately interprets scene geometry through our targeted view depth estimation design, allowing us to not only precisely segment objects but also render novel view images more realistically. This confirms that our depth-guided sampling technique effectively captures the correct geometry and that the sampled point features are appropriately aggregated from features of the Semantic Geo-Reasoner.

\subsubsection{Ablation Studies} 
To further analyze the effectiveness of our designed modules, we conduct ablation studies on ScanNet, as shown in Table~\ref{tab:ablation_board}. Baseline model A employs our Semantic Geo-Reasoner $G_\theta$ and volume renderer $R_\theta$ to create the target view image $I_T$, using uniform sampling instead of depth-guided sampling $x_{1:N}$. Also, a pre-trained 2D segmentor~\cite{ronneberger2015u} is applied on the rendered image $I_T$ to produce $S_T$ without our semantic renderer $P_\theta$. Model B enhances visual quality by replacing uniform sampling with depth-guided sampling. We further replace the 2D segmentor with our semantic renderer $P_{\theta}$ using uniform sampling for semantic rendering in model C, illustrating mutual benefits for both tasks. Our full model in the last row uses depth-guided sampling exclusively for surface points in the semantic renderer, achieving optimal results. This verifies the success of our proposed modules and sampling strategies.
\vspace{-1.5mm}
\subsubsection{Sampling Efficiency}
Another benefit of our depth-guided sampling is that the performance of our GSNeRF is less sensitive to the number of sampled points along each ray. In Fig.~\ref{fig:points}, we conduct experiments of reducing the number of sampling points $N$, showing PSNR of the synthesized novel view images as evaluation. We find that by employing depth-guided sampling, our method accurately preserves visual details with less noise (compared to Neuray and S-Ray with 128 sampled points along each ray). It can be seen that GeoNeRF + semhead shows a catastrophic drop in PSNR with fewer points, while ours leveraging depth-guided sampling maintains satisfactory visual quality even with a significantly reduced number of sampled points (\eg, when sampling only 4 points along a ray, the PSNR remains near 28). This result further attests to the success of our design in estimating target depth and implementing the corresponding sampling strategy with better novel view synthesis performance.

\section{Conclusion and Limitations}
\paragraph{Conclusion}
In this paper,  we proposed Generalizable Semantic Neural Raidance Fields (GSNeRF) to achieve generalized novel view synthesis and semantic segmentation. our GSNeRF is trained to derive visual features and perform depth map prediction of each source view so that the depth map for a novel target view can be estimated. With such target-view depth information observed, the associated RGB image and semantic segmentation can be jointly produced via depth-guide rendering. In our experiments, we quantitatively and qualitatively confirm that our GSNeRF performs favorably against existing generalizable semantic-aware NeRF methods on both real-world and synthetic datasets.

\paragraph{Limitations}
Our GSNeRF is proposed for novel-view scene synthesis and understanding in a generalized setting. Unlike studies like~\cite{hu2023sherf, liao2023high, weng2023zeroavatar, gao2020portrait} which synthesize novel-view images for particular objects like human or face from a single image, our method utilizes self-supervised loss in Eq.~\ref{eq:ssloss} for observing cross-view depth consistency. Therefore, extensions of our work to single-image generalizable NeRF for specific 3D objects would be among our future research directions.

\paragraph{Acknowledgements}
This work is supported in part by the National Science and Technology Council under grant 112-2634-F-002-007. We also thank to National Center for High-performance Computing (NCHC) for providing computational and storage resources.

{    \small
    \bibliographystyle{ieeenat_fullname}
    \bibliography{main}

\begin{thebibliography}{44}
\providecommand{\natexlab}[1]{#1}
\providecommand{\url}[1]{\texttt{#1}}
\expandafter\ifx\csname urlstyle\endcsname\relax
  \providecommand{\doi}[1]{doi: #1}\else
  \providecommand{\doi}{doi: \begingroup \urlstyle{rm}\Url}\fi

\bibitem[Barron et~al.(2021)Barron, Mildenhall, Tancik, Hedman, Martin-Brualla, and Srinivasan]{barron2021mip}
Jonathan~T Barron, Ben Mildenhall, Matthew Tancik, Peter Hedman, Ricardo Martin-Brualla, and Pratul~P Srinivasan.
\newblock Mip-nerf: A multiscale representation for anti-aliasing neural radiance fields.
\newblock In \emph{Proceedings of the IEEE/CVF International Conference on Computer Vision}, pages 5855--5864, 2021.

\bibitem[Chang et~al.(2022)Chang, Bo{\v{z}}i{\v{c}}, Zhang, Yan, Chen, S{\"u}sstrunk, and Nie{\ss}ner]{chang2022rc}
Di Chang, Alja{\v{z}} Bo{\v{z}}i{\v{c}}, Tong Zhang, Qingsong Yan, Yingcong Chen, Sabine S{\"u}sstrunk, and Matthias Nie{\ss}ner.
\newblock Rc-mvsnet: unsupervised multi-view stereo with neural rendering.
\newblock In \emph{European Conference on Computer Vision}, pages 665--680. Springer, 2022.

\bibitem[Chen et~al.(2021)Chen, Xu, Zhao, Zhang, Xiang, Yu, and Su]{chen2021mvsnerf}
Anpei Chen, Zexiang Xu, Fuqiang Zhao, Xiaoshuai Zhang, Fanbo Xiang, Jingyi Yu, and Hao Su.
\newblock Mvsnerf: Fast generalizable radiance field reconstruction from multi-view stereo.
\newblock In \emph{Proceedings of the IEEE/CVF International Conference on Computer Vision}, pages 14124--14133, 2021.

\bibitem[Chen et~al.(2022)Chen, Xu, Geiger, Yu, and Su]{chen2022tensorf}
Anpei Chen, Zexiang Xu, Andreas Geiger, Jingyi Yu, and Hao Su.
\newblock Tensorf: Tensorial radiance fields.
\newblock In \emph{European Conference on Computer Vision}, pages 333--350. Springer, 2022.

\bibitem[Dai et~al.(2017)Dai, Chang, Savva, Halber, Funkhouser, and Nie{\ss}ner]{dai2017scannet}
Angela Dai, Angel~X Chang, Manolis Savva, Maciej Halber, Thomas Funkhouser, and Matthias Nie{\ss}ner.
\newblock Scannet: Richly-annotated 3d reconstructions of indoor scenes.
\newblock In \emph{Proceedings of the IEEE conference on computer vision and pattern recognition}, pages 5828--5839, 2017.

\bibitem[Dai et~al.(2019)Dai, Zhu, Rao, and Li]{dai2019mvs2}
Yuchao Dai, Zhidong Zhu, Zhibo Rao, and Bo Li.
\newblock Mvs2: Deep unsupervised multi-view stereo with multi-view symmetry.
\newblock In \emph{2019 International Conference on 3D Vision (3DV)}, pages 1--8. Ieee, 2019.

\bibitem[Deng et~al.(2022)Deng, Liu, Zhu, and Ramanan]{deng2022depth}
Kangle Deng, Andrew Liu, Jun-Yan Zhu, and Deva Ramanan.
\newblock Depth-supervised nerf: Fewer views and faster training for free.
\newblock In \emph{Proceedings of the IEEE/CVF Conference on Computer Vision and Pattern Recognition}, pages 12882--12891, 2022.

\bibitem[Fridovich-Keil et~al.(2022)Fridovich-Keil, Yu, Tancik, Chen, Recht, and Kanazawa]{fridovich2022plenoxels}
Sara Fridovich-Keil, Alex Yu, Matthew Tancik, Qinhong Chen, Benjamin Recht, and Angjoo Kanazawa.
\newblock Plenoxels: Radiance fields without neural networks.
\newblock In \emph{Proceedings of the IEEE/CVF Conference on Computer Vision and Pattern Recognition}, pages 5501--5510, 2022.

\bibitem[Fu et~al.(2022)Fu, Zhang, Chen, Lu, Zhu, Zhou, Geiger, and Liao]{fu2022panoptic}
Xiao Fu, Shangzhan Zhang, Tianrun Chen, Yichong Lu, Lanyun Zhu, Xiaowei Zhou, Andreas Geiger, and Yiyi Liao.
\newblock Panoptic nerf: 3d-to-2d label transfer for panoptic urban scene segmentation.
\newblock In \emph{2022 International Conference on 3D Vision (3DV)}, pages 1--11. IEEE, 2022.

\bibitem[Gao et~al.(2020)Gao, Shih, Lai, Liang, and Huang]{gao2020portrait}
Chen Gao, Yichang Shih, Wei-Sheng Lai, Chia-Kai Liang, and Jia-Bin Huang.
\newblock Portrait neural radiance fields from a single image.
\newblock \emph{arXiv preprint arXiv:2012.05903}, 2020.

\bibitem[Garg et~al.(2020)Garg, S{\"u}nderhauf, Dayoub, Morrison, Cosgun, Carneiro, Wu, Chin, Reid, Gould, et~al.]{garg2020semantics}
Sourav Garg, Niko S{\"u}nderhauf, Feras Dayoub, Douglas Morrison, Akansel Cosgun, Gustavo Carneiro, Qi Wu, Tat-Jun Chin, Ian Reid, Stephen Gould, et~al.
\newblock Semantics for robotic mapping, perception and interaction: A survey.
\newblock \emph{Foundations and Trends{\textregistered} in Robotics}, 8\penalty0 (1--2):\penalty0 1--224, 2020.

\bibitem[Girshick(2015)]{girshick2015fast}
Ross Girshick.
\newblock Fast r-cnn.
\newblock In \emph{Proceedings of the IEEE international conference on computer vision}, pages 1440--1448, 2015.

\bibitem[Gu et~al.(2020)Gu, Fan, Zhu, Dai, Tan, and Tan]{gu2020cascade}
Xiaodong Gu, Zhiwen Fan, Siyu Zhu, Zuozhuo Dai, Feitong Tan, and Ping Tan.
\newblock Cascade cost volume for high-resolution multi-view stereo and stereo matching.
\newblock In \emph{Proceedings of the IEEE/CVF conference on computer vision and pattern recognition}, pages 2495--2504, 2020.

\bibitem[Hu et~al.(2023)Hu, Hong, Pan, Mei, Yang, and Liu]{hu2023sherf}
Shoukang Hu, Fangzhou Hong, Liang Pan, Haiyi Mei, Lei Yang, and Ziwei Liu.
\newblock Sherf: Generalizable human nerf from a single image.
\newblock \emph{arXiv preprint arXiv:2303.12791}, 2023.

\bibitem[Huang et~al.(2021)Huang, Yi, Huang, He, Liu, and Liu]{huang2021m3vsnet}
Baichuan Huang, Hongwei Yi, Can Huang, Yijia He, Jingbin Liu, and Xiao Liu.
\newblock M3vsnet: Unsupervised multi-metric multi-view stereo network.
\newblock In \emph{2021 IEEE International Conference on Image Processing (ICIP)}, pages 3163--3167. IEEE, 2021.

\bibitem[Johari et~al.(2022)Johari, Lepoittevin, and Fleuret]{johari2022geonerf}
Mohammad~Mahdi Johari, Yann Lepoittevin, and Fran{\c{c}}ois Fleuret.
\newblock Geonerf: Generalizing nerf with geometry priors.
\newblock In \emph{Proceedings of the IEEE/CVF Conference on Computer Vision and Pattern Recognition}, pages 18365--18375, 2022.

\bibitem[Kingma and Ba(2014)]{kingma2014adam}
Diederik~P Kingma and Jimmy Ba.
\newblock Adam: A method for stochastic optimization.
\newblock \emph{arXiv preprint arXiv:1412.6980}, 2014.

\bibitem[Kundu et~al.(2022)Kundu, Genova, Yin, Fathi, Pantofaru, Guibas, Tagliasacchi, Dellaert, and Funkhouser]{kundu2022panoptic}
Abhijit Kundu, Kyle Genova, Xiaoqi Yin, Alireza Fathi, Caroline Pantofaru, Leonidas~J Guibas, Andrea Tagliasacchi, Frank Dellaert, and Thomas Funkhouser.
\newblock Panoptic neural fields: A semantic object-aware neural scene representation.
\newblock In \emph{Proceedings of the IEEE/CVF Conference on Computer Vision and Pattern Recognition}, pages 12871--12881, 2022.

\bibitem[Liao et~al.(2023)Liao, Zhang, Xiu, Yi, Liu, Qi, Zhang, Wang, Zhu, and Lei]{liao2023high}
Tingting Liao, Xiaomei Zhang, Yuliang Xiu, Hongwei Yi, Xudong Liu, Guo-Jun Qi, Yong Zhang, Xuan Wang, Xiangyu Zhu, and Zhen Lei.
\newblock High-fidelity clothed avatar reconstruction from a single image.
\newblock In \emph{Proceedings of the IEEE/CVF Conference on Computer Vision and Pattern Recognition}, pages 8662--8672, 2023.

\bibitem[Liu et~al.(2023)Liu, Zhang, Zheng, and Duan]{liu2023semantic}
Fangfu Liu, Chubin Zhang, Yu Zheng, and Yueqi Duan.
\newblock Semantic ray: Learning a generalizable semantic field with cross-reprojection attention.
\newblock In \emph{CVPR}, pages 17386--17396, 2023.

\bibitem[Liu et~al.(2022)Liu, Peng, Liu, Wang, Wang, Theobalt, Zhou, and Wang]{liu2022neural}
Yuan Liu, Sida Peng, Lingjie Liu, Qianqian Wang, Peng Wang, Christian Theobalt, Xiaowei Zhou, and Wenping Wang.
\newblock Neural rays for occlusion-aware image-based rendering.
\newblock In \emph{Proceedings of the IEEE/CVF Conference on Computer Vision and Pattern Recognition}, pages 7824--7833, 2022.

\bibitem[Martin-Brualla et~al.(2021)Martin-Brualla, Radwan, Sajjadi, Barron, Dosovitskiy, and Duckworth]{martin2021nerf}
Ricardo Martin-Brualla, Noha Radwan, Mehdi~SM Sajjadi, Jonathan~T Barron, Alexey Dosovitskiy, and Daniel Duckworth.
\newblock Nerf in the wild: Neural radiance fields for unconstrained photo collections.
\newblock In \emph{Proceedings of the IEEE/CVF Conference on Computer Vision and Pattern Recognition}, pages 7210--7219, 2021.

\bibitem[Mildenhall et~al.(2021)Mildenhall, Srinivasan, Tancik, Barron, Ramamoorthi, and Ng]{mildenhall2021nerf}
Ben Mildenhall, Pratul~P Srinivasan, Matthew Tancik, Jonathan~T Barron, Ravi Ramamoorthi, and Ren Ng.
\newblock Nerf: Representing scenes as neural radiance fields for view synthesis.
\newblock \emph{Communications of the ACM}, 65\penalty0 (1):\penalty0 99--106, 2021.

\bibitem[M{\"u}ller et~al.(2022)M{\"u}ller, Evans, Schied, and Keller]{muller2022instant}
Thomas M{\"u}ller, Alex Evans, Christoph Schied, and Alexander Keller.
\newblock Instant neural graphics primitives with a multiresolution hash encoding.
\newblock \emph{ACM Transactions on Graphics (ToG)}, 41\penalty0 (4):\penalty0 1--15, 2022.

\bibitem[Paszke et~al.(2019)Paszke, Gross, Massa, Lerer, Bradbury, Chanan, Killeen, Lin, Gimelshein, Antiga, et~al.]{paszke2019pytorch}
Adam Paszke, Sam Gross, Francisco Massa, Adam Lerer, James Bradbury, Gregory Chanan, Trevor Killeen, Zeming Lin, Natalia Gimelshein, Luca Antiga, et~al.
\newblock Pytorch: An imperative style, high-performance deep learning library.
\newblock \emph{Advances in neural information processing systems}, 32, 2019.

\bibitem[Reiser et~al.(2021)Reiser, Peng, Liao, and Geiger]{reiser2021kilonerf}
Christian Reiser, Songyou Peng, Yiyi Liao, and Andreas Geiger.
\newblock Kilonerf: Speeding up neural radiance fields with thousands of tiny mlps.
\newblock In \emph{Proceedings of the IEEE/CVF International Conference on Computer Vision}, pages 14335--14345, 2021.

\bibitem[Roessle et~al.(2022)Roessle, Barron, Mildenhall, Srinivasan, and Nie{\ss}ner]{roessle2022dense}
Barbara Roessle, Jonathan~T Barron, Ben Mildenhall, Pratul~P Srinivasan, and Matthias Nie{\ss}ner.
\newblock Dense depth priors for neural radiance fields from sparse input views.
\newblock In \emph{Proceedings of the IEEE/CVF Conference on Computer Vision and Pattern Recognition}, pages 12892--12901, 2022.

\bibitem[Ronneberger et~al.(2015)Ronneberger, Fischer, and Brox]{ronneberger2015u}
Olaf Ronneberger, Philipp Fischer, and Thomas Brox.
\newblock U-net: Convolutional networks for biomedical image segmentation.
\newblock In \emph{Medical Image Computing and Computer-Assisted Intervention--MICCAI 2015: 18th International Conference, Munich, Germany, October 5-9, 2015, Proceedings, Part III 18}, pages 234--241. Springer, 2015.

\bibitem[Siddiqui et~al.(2023)Siddiqui, Porzi, Bul{\`o}, M{\"u}ller, Nie{\ss}ner, Dai, and Kontschieder]{siddiqui2023panoptic}
Yawar Siddiqui, Lorenzo Porzi, Samuel~Rota Bul{\`o}, Norman M{\"u}ller, Matthias Nie{\ss}ner, Angela Dai, and Peter Kontschieder.
\newblock Panoptic lifting for 3d scene understanding with neural fields.
\newblock In \emph{Proceedings of the IEEE/CVF Conference on Computer Vision and Pattern Recognition}, pages 9043--9052, 2023.

\bibitem[Straub et~al.(2019)Straub, Whelan, Ma, Chen, Wijmans, Green, Engel, Mur-Artal, Ren, Verma, et~al.]{straub2019replica}
Julian Straub, Thomas Whelan, Lingni Ma, Yufan Chen, Erik Wijmans, Simon Green, Jakob~J Engel, Raul Mur-Artal, Carl Ren, Shobhit Verma, et~al.
\newblock The replica dataset: A digital replica of indoor spaces.
\newblock \emph{arXiv preprint arXiv:1906.05797}, 2019.

\bibitem[Suhail et~al.(2022)Suhail, Esteves, Sigal, and Makadia]{suhail2022generalizable}
Mohammed Suhail, Carlos Esteves, Leonid Sigal, and Ameesh Makadia.
\newblock Generalizable patch-based neural rendering.
\newblock In \emph{European Conference on Computer Vision}, pages 156--174. Springer, 2022.

\bibitem[Sun et~al.(2022)Sun, Sun, and Chen]{sun2022direct}
Cheng Sun, Min Sun, and Hwann-Tzong Chen.
\newblock Direct voxel grid optimization: Super-fast convergence for radiance fields reconstruction.
\newblock In \emph{Proceedings of the IEEE/CVF Conference on Computer Vision and Pattern Recognition}, pages 5459--5469, 2022.

\bibitem[Trevithick and Yang(2021)]{trevithick2021grf}
Alex Trevithick and Bo Yang.
\newblock Grf: Learning a general radiance field for 3d representation and rendering.
\newblock In \emph{Proceedings of the IEEE/CVF International Conference on Computer Vision}, pages 15182--15192, 2021.

\bibitem[Vora et~al.(2021)Vora, Radwan, Greff, Meyer, Genova, Sajjadi, Pot, Tagliasacchi, and Duckworth]{vora2021nesf}
Suhani Vora, Noha Radwan, Klaus Greff, Henning Meyer, Kyle Genova, Mehdi~SM Sajjadi, Etienne Pot, Andrea Tagliasacchi, and Daniel Duckworth.
\newblock Nesf: Neural semantic fields for generalizable semantic segmentation of 3d scenes.
\newblock \emph{arXiv preprint arXiv:2111.13260}, 2021.

\bibitem[Wang et~al.(2022{\natexlab{a}})Wang, Chen, and Yang]{wang2022dm}
Bing Wang, Lu Chen, and Bo Yang.
\newblock Dm-nerf: 3d scene geometry decomposition and manipulation from 2d images.
\newblock \emph{arXiv preprint arXiv:2208.07227}, 2022{\natexlab{a}}.

\bibitem[Wang et~al.(2022{\natexlab{b}})Wang, Chen, Chen, Venugopalan, Wang, et~al.]{wang2022attention}
Peihao Wang, Xuxi Chen, Tianlong Chen, Subhashini Venugopalan, Zhangyang Wang, et~al.
\newblock Is attention all nerf needs?
\newblock \emph{arXiv preprint arXiv:2207.13298}, 2022{\natexlab{b}}.

\bibitem[Wang et~al.(2021)Wang, Wang, Genova, Srinivasan, Zhou, Barron, Martin-Brualla, Snavely, and Funkhouser]{wang2021ibrnet}
Qianqian Wang, Zhicheng Wang, Kyle Genova, Pratul~P Srinivasan, Howard Zhou, Jonathan~T Barron, Ricardo Martin-Brualla, Noah Snavely, and Thomas Funkhouser.
\newblock Ibrnet: Learning multi-view image-based rendering.
\newblock In \emph{Proceedings of the IEEE/CVF Conference on Computer Vision and Pattern Recognition}, pages 4690--4699, 2021.

\bibitem[Wang et~al.(2004)Wang, Bovik, Sheikh, and Simoncelli]{wang2004image}
Zhou Wang, Alan~C Bovik, Hamid~R Sheikh, and Eero~P Simoncelli.
\newblock Image quality assessment: from error visibility to structural similarity.
\newblock \emph{IEEE transactions on image processing}, 13\penalty0 (4):\penalty0 600--612, 2004.

\bibitem[Weng et~al.(2023)Weng, Wang, and Yeung]{weng2023zeroavatar}
Zhenzhen Weng, Zeyu Wang, and Serena Yeung.
\newblock Zeroavatar: Zero-shot 3d avatar generation from a single image.
\newblock \emph{arXiv preprint arXiv:2305.16411}, 2023.

\bibitem[Yao et~al.(2018)Yao, Luo, Li, Fang, and Quan]{yao2018mvsnet}
Yao Yao, Zixin Luo, Shiwei Li, Tian Fang, and Long Quan.
\newblock Mvsnet: Depth inference for unstructured multi-view stereo.
\newblock In \emph{Proceedings of the European conference on computer vision (ECCV)}, pages 767--783, 2018.

\bibitem[Yu et~al.(2021{\natexlab{a}})Yu, Li, Tancik, Li, Ng, and Kanazawa]{yu2021plenoctrees}
Alex Yu, Ruilong Li, Matthew Tancik, Hao Li, Ren Ng, and Angjoo Kanazawa.
\newblock Plenoctrees for real-time rendering of neural radiance fields.
\newblock In \emph{Proceedings of the IEEE/CVF International Conference on Computer Vision}, pages 5752--5761, 2021{\natexlab{a}}.

\bibitem[Yu et~al.(2021{\natexlab{b}})Yu, Ye, Tancik, and Kanazawa]{yu2021pixelnerf}
Alex Yu, Vickie Ye, Matthew Tancik, and Angjoo Kanazawa.
\newblock pixelnerf: Neural radiance fields from one or few images.
\newblock In \emph{Proceedings of the IEEE/CVF Conference on Computer Vision and Pattern Recognition}, pages 4578--4587, 2021{\natexlab{b}}.

\bibitem[Zhang et~al.(2018)Zhang, Isola, Efros, Shechtman, and Wang]{zhang2018unreasonable}
Richard Zhang, Phillip Isola, Alexei~A Efros, Eli Shechtman, and Oliver Wang.
\newblock The unreasonable effectiveness of deep features as a perceptual metric.
\newblock In \emph{Proceedings of the IEEE conference on computer vision and pattern recognition}, pages 586--595, 2018.

\bibitem[Zhi et~al.(2021)Zhi, Laidlow, Leutenegger, and Davison]{zhi2021place}
Shuaifeng Zhi, Tristan Laidlow, Stefan Leutenegger, and Andrew~J Davison.
\newblock In-place scene labelling and understanding with implicit scene representation.
\newblock In \emph{Proceedings of the IEEE/CVF International Conference on Computer Vision}, pages 15838--15847, 2021.

\end{thebibliography}
}
\clearpage
\setcounter{page}{1}
\maketitlesupplementary

\appendix
\section{Additional implementation Details}

\subsection{Self-Supervised Depth Loss}
Our method generalizes well to unseen scenes, either with GT depth for supervision or with self-supervised depth regularization.
As detailed in Section~\textcolor{red}{4.2} of our main paper, we ensure the accuracy of $D_{1:K}$ even when GT depths are unavailable through applying a self-supervised loss function $\mathcal{L}_{ssl}$~\cite{dai2019mvs2,huang2021m3vsnet,chang2022rc} to regularize our depth predictions. By focusing on the cross-view depth consistency among all source views, we are able to regularize our depth predictions effectively. This loss function (Eq.~\textcolor{red}{7} in our main paper) is defined as:
\begin{equation} \label{eq:ssloss_sup}
\mathcal{L}_{ssl} = \lambda_1\mathcal{L}_{RC} + \lambda_2\mathcal{L}_{SSIM} + \lambda_3\mathcal{L}_{Smooth},
\end{equation}
where $\mathcal{L}_{RC}$ calculates the mean square error between a source image $I_k$ and the reconstructed image $I'_k$ obtained by warping other source images using their predicted depth maps for all source images $I_{1:K}$. $\mathcal{L}_{SSIM}$ measures the structural similarity between $I'_{1:K}$ and $I_{1:K}$, and $\mathcal{L}_{Smooth}$ ensures the smoothness of all depth prediction $D_{1:K}$ by penalizing large variations in the depth values between neighboring pixels. Following~\cite{chang2022rc}, the hyperparameters are set at $\lambda_1 = 1$, $\lambda_2 = 0.2$, and $\lambda_3 = 0.0067$.

\subsection{Target View Depth Estimation}
As mentioned in Section \textcolor{red}{4.2} of our main paper, we estimate target view depth $D_T$ from the depth prediction of each source image $D_{1:K}$, corresponding source view camera poses $\xi_{1:K}$ and the target view camera pose $\xi_T$. To be more specific, $D_T$ is estimated by projecting the pixels of each depth map into 3D space and then reprojecting them back to the target view. The pseudo-code of the target view depth estimation is described in Algorithm~\ref{alg:algorithm}. 

\begin{algorithm}[tb]
\caption{Estimation of Depth in Target View}
\label{alg:algorithm}
\textbf{Input}: Depth predictions of each source view $D_{1:K}$, camera pose of each source view $\xi_{1:K}$, target camera pose $\xi_T$\\
\textbf{Data}: Image size: (H, W), camera pose of the world coordinate $\xi_w$
\begin{flushleft}
\textbf{Output}: Target view depth estimation $D_T$
\end{flushleft}
\begin{algorithmic}[1] 
\STATE $A$ $\leftarrow$ empty array()
\FOR{$k$ = $1,...,K$}
    \STATE $g$ $\leftarrow$ meshgrid(H, W)
    \STATE Project $g$ into the coordinate system defined by $\xi_k$
    \STATE Multiply $g$ by the corresponding depth prediction $D_k$
    \STATE $g$ $\leftarrow$ \textbf{Transform}($g$, $\xi_k$, $\xi_{w}$)
    \STATE Append $g$ to the array $A$
\ENDFOR
\STATE $A$ $\leftarrow$ \textbf{Transform}($A$, $\xi_{w}$, $\xi_T$)
\STATE Reproject $A$ onto the $\xi_T$ image plane
\STATE $Z$ $\leftarrow$ the third element (Z-axis) of points $A$ 
\STATE $A'$ $\leftarrow$ round the first two elements of $A$ to integer values
\STATE $W$ $\leftarrow$ The first two elements of ($A'$ - $A$)
\STATE Weight and normalize $Z$ using weight $W$
\STATE Set the depth of target view $D_T$ to $Z$ based on the index of the first two elements of $A'$
\RETURN Estimated depth of target view $D_T$
\STATE
\STATE \textit{/* Function */}
\STATE \textbf{Transform}(point, $\xi_1$, $\xi_2$):
\RETURN transform point from coordinate $\xi_1$ to $\xi_2$
\end{algorithmic}
\end{algorithm}

\subsection{Masking Unrelated Features for Depth-Guided Visual Rendering}
In Section \textcolor{red}{4.2.2} of the main paper, we utilize occlusion-aware masks represented by $M_n$ for volume rendering in Eq.~\textcolor{red}{11} and $M$ for semantic rendering in Eq.~\textcolor{red}{12}. These masks selectively exclude irrelevant features of points located behind object surfaces. It is worth noting that, when computing the global features $f_{n,0}$ in Eq.~\textcolor{red}{10} and $f_0$ and in Eq.~\textcolor{red}{14}, the mean and variance are also calculated as masked mean and variance using $M_n$ and $M$, respectively, ensuring the exclusion of unrelated information.

\subsection{Training Strategy for Depth-Guided Volume Rendering}
During the volume rendering process in Section~\textcolor{red}{4.2}, we sample points along the ray based on the estimated target view depth map $D_T$. However, in the early phase of our training, such estimation might not be accurate. Therefore, we employ a mix of uniform and depth-guided sampling, using half the points for each for the first 125K training steps and then switching to all depth-guided for the rest 125K steps. This approach stabilizes our volume rendering process and makes sure that our GSNeRF predicts accurate colors for each pixel of the target view image.

\renewcommand{\arraystretch}{1.3}
\begin{table}[t]
\resizebox{1\linewidth}{!}{%
\begin{tabular}{l|cl|ccc}
\toprule
\multirow{2}{*}{Finetuned Method} & \multicolumn{2}{c|}{GT Depth} & \multicolumn{3}{c}{ScanNet} \\ \cmidrule{2-6} 
 & \multicolumn{2}{c|}{Train / Test} & mIoU & acc. / class acc. & PSNR \\ \midrule
S-Ray & \multicolumn{2}{c|}{\cmark\ /\ \cmark} & 92.4 & 98.2 / 93.8 & 27.67 \\
Ours & \multicolumn{2}{c|}{\cmark\ /\ \,\,\,\,\,\,} & \textbf{93.9} & \textbf{99.1} / \textbf{98.4} & \textbf{31.70} \\ \midrule
S-Ray & \multicolumn{2}{c|}{/\,\,} & 91.6 & 97.3 / 92.2 & 27.31 \\
Ours & \multicolumn{2}{c|}{/\,\,} & \textbf{93.2} & \textbf{98.2} / \textbf{96.8} & \textbf{30.89} \\
\bottomrule
\end{tabular}}
\caption{\textbf{Results of finetuning on unseen scenes of ScanNet. } Note that methods in the first two rows take GT depth during training, while S-Ray additionally requires such inputs during testing. The methods in the last two rows do not have access to GT depth during training/testing.}
\label{finetune_table}
\end{table}
\begin{figure}[t]
\resizebox{\linewidth}{!}{%

\begin{tabular}{ccc}
 
 {\fontsize{30}{36}\selectfont S-Ray} & 
 {\fontsize{30}{36}\selectfont GSNeRF (Ours)}&
 {\fontsize{30}{36}\selectfont GT} \\

\includegraphics[]{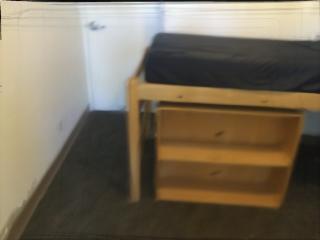} &
\includegraphics[]{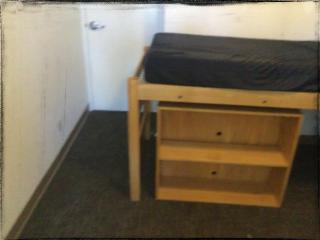}  &
\includegraphics[]{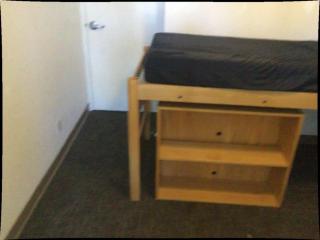}\\

\includegraphics[]{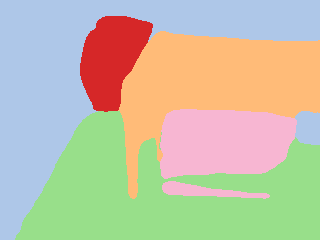} &
\includegraphics[]{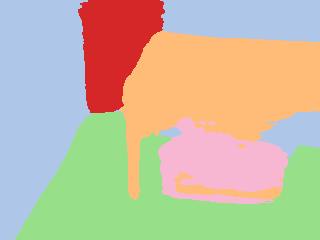}  &
\includegraphics[]{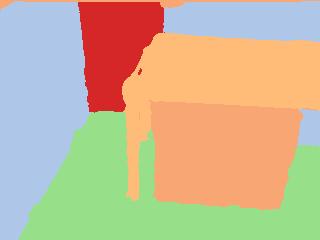}  \\
 
\end{tabular}}

\caption{\textbf{Qualitative results of finetuning on ScanNet.} Unlike our GSNeRF, S-Ray fails to capture the semantic contour of the door (in red) at the upper-left corner.
}
\label{fig:fintune}
\vspace{-0mm}
\end{figure}

\subsection{More Training Details}
Given a set of multi-view images of a scene, we select training pairs of source view and target view by first randomly selecting a target view, and sampling $K$ nearby yet sparse views as source views, following the setting of S-Ray~\cite{liu2023semantic} and NeuRay~\cite{liu2022neural}. We implement our model using PyTorch~\cite{paszke2019pytorch} and train it end-to-end on a single RTX3090Ti GPU with 24G memory. Notably, we did not utilize any pre-trained weights. The batch size of rays is set to 1024 and our model is trained for 250k steps using Adam optimizer~\cite{kingma2014adam} with an initial learning rate of 5e-4 decaying to 1e-5.

\subsection{Evaluation Metrics}
To evaluate the effectiveness of our method, we examine both semantic performance and visual quality through various metrics in Table~\textcolor{red}{2} of our main paper. We measure the semantic capabilities of our approach using the mean Intersection over Union (mIoU), class average pixel accuracy, and total pixel accuracy. These metrics provide a comprehensive evaluation of how accurately our method is able to recognize and delineate semantic objects within the scene. For evaluating the visual fidelity of the synthesized images, we employ peak signal-to-noise ratio (PSNR), the structural similarity index measure (SSIM)~\cite{wang2004image}, and learned perceptual image patch similarity (LPIPS)~\cite{zhang2018unreasonable}. These metrics collectively assess the clarity, structural integrity, and perceptual resemblance of the rendered images as compared to the ground truth. 

\begin{table}[t]
\centering
\resizebox{0.65\linewidth}{!}{%
\begin{tabular}{ccccc}
\toprule
K & mIoU & acc. & class acc. & PSNR \\ \midrule
4 & 48.70 & 72.71 & 57.97 & 31.02 \\
6 & 51.61 & 73.92 & 59.45 & 30.96 \\
8 & \textbf{58.30} & \textbf{79.79} & \textbf{65.93} & \textbf{31.33} \\ \bottomrule
\end{tabular}}
\caption{\textbf{Comparisons of different numbers of source-view images on ScanNet} We show the quantitative results of our method, given K = 4, 6, or 8 input views. The testing scene is not seen during training (i.e., the generalized setting).}
\label{different}
\end{table}
\begin{figure}[t]
\resizebox{\linewidth}{!}{%

\begin{tabular}{ccc}

 {\fontsize{30}{36}\selectfont GeoNeRF + semhead} & 
 {\fontsize{30}{36}\selectfont GSNeRF (Ours)} &
{\fontsize{30}{36}\selectfont GT} \\

\includegraphics[]{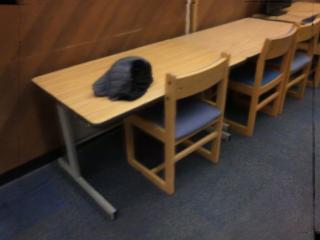} &
\includegraphics[]{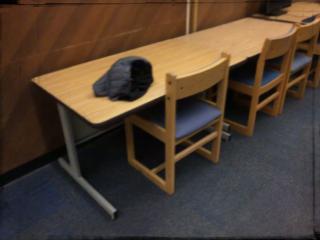} &
\includegraphics[]{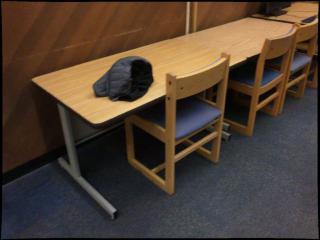} \\

\includegraphics[]{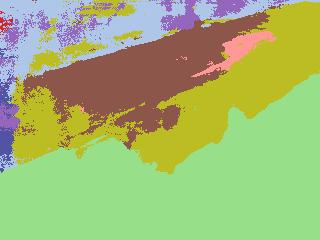} &
\includegraphics[]{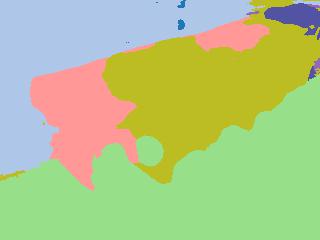} & 
\includegraphics[]{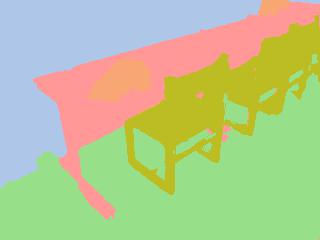}  \\
 
\end{tabular}}

\caption{\textbf{Qualitative comparisons with GeoNeRF with semhead on ScanNet.} While the difference between the visual quality of rendering images (in the first row) is not remarkable, improved semantic segmentation can be observed for our GSNeRF (in the second row).
}
\label{fig:geonerf_compare}
\vspace{-0mm}
\end{figure}

\section{Additional Experiments and Analysis}
\subsection{Finetuning on Unseen Scenes}
To enhance the completeness of our method, we adopt the fine-tuning setting in S-Ray~\cite{liu2023semantic}. Specifically, we fine-tune our generalized model for a limited number of steps, 5k steps, on each unseen scene before evaluation. 

\begin{figure*}[t]
\Huge
\resizebox{\textwidth}{!}{%

\begin{tabular}{cc|c|cc|c}
 {\fontsize{40}{48}\selectfont S-Ray} & 
 {\fontsize{40}{48}\selectfont GSNeRF (Ours)} &
 {\fontsize{40}{48}\selectfont GT} &
 {\fontsize{40}{48}\selectfont S-Ray} &
 {\fontsize{40}{48}\selectfont GSNeRF (Ours)} & 
 {\fontsize{40}{48}\selectfont GT} \\
 
\includegraphics[]{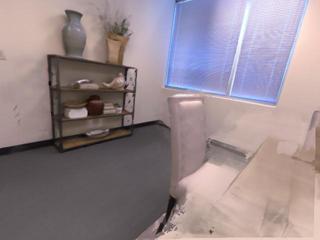} &
\includegraphics[]{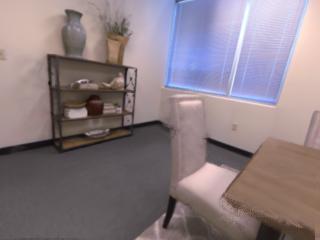} &
\includegraphics[]{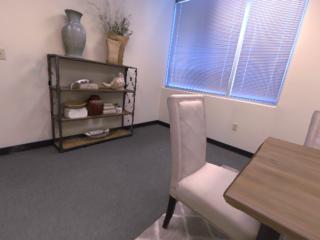} &
\includegraphics[]{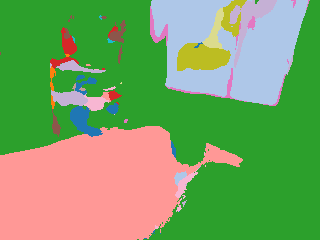} &
\includegraphics[]{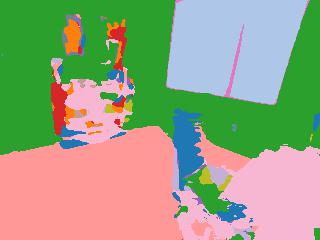} & 
\includegraphics[]{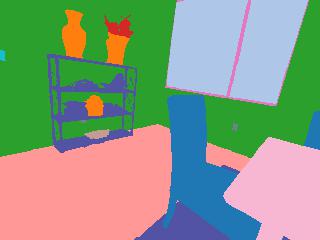} \\

\includegraphics[]{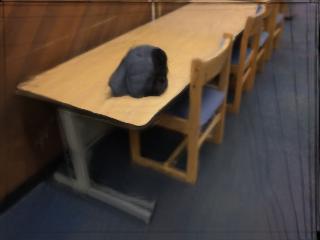} &
\includegraphics[]{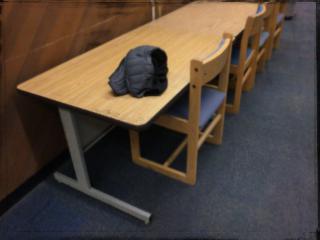} &
\includegraphics[]{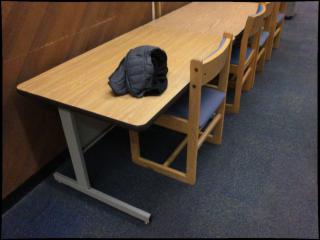} &
\includegraphics[]{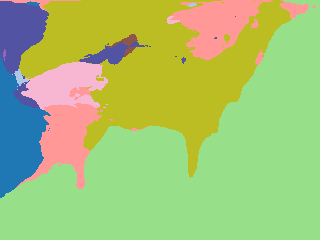} &
\includegraphics[]{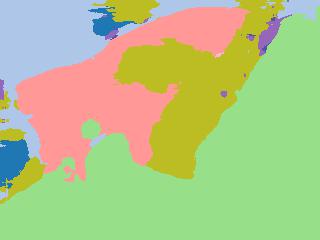} &
\includegraphics[]{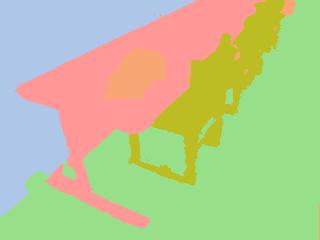}  \\

\includegraphics[]{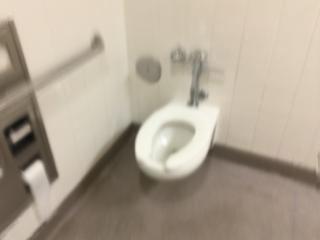} &
\includegraphics[]{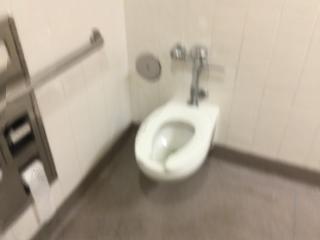} &
\includegraphics[]{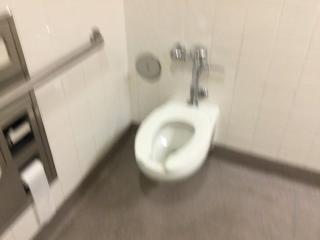} &
\includegraphics[]{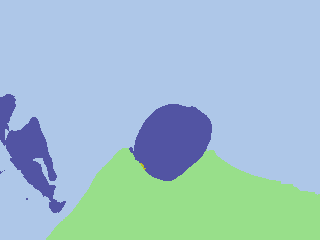} &
\includegraphics[]{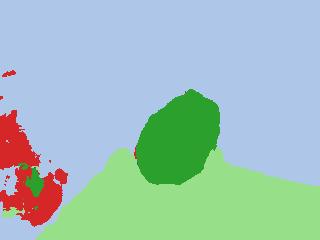} &
\includegraphics[]{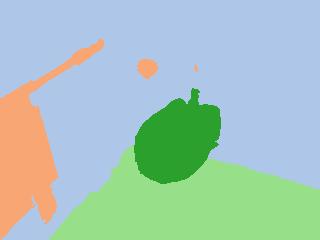}  \\

\end{tabular}}

\caption{\textbf{More qualitative evaluation.} We compare the visual quality of the rendered novel view images (the first three columns) and semantic segmentation maps (the last three columns) with S-Ray~\cite{liu2023semantic}. Our method shows clearer image rendering quality and better semantic segmentation results.}
\label{fig:quality_image}
\end{figure*}

Table~\ref{finetune_table} and Fig.~\ref{fig:fintune} show the quantitative and quantitative results of our model finetuning 5k steps on ScanNet~\cite{dai2017scannet}. We observe that by adopting our designed Semantic Geo-Reasoning and Depth-Guided Visual Rendering, our method preserved better rendering quality in the finetuning setting. We further include .mp4 files of trajectories provided in ScanNet for better visualization. (i.e., \texttt{finetune\_compare.mp4} shows qualitative comparison of our method compare with S-Ray~\cite{liu2023semantic} under fintuning setting. \texttt{finetune\_ours.mp4} shows our results with predicting depth map and RGB rendering error map.)

\subsection{Observations on Different Number of Source Views}
Even though we followed S-Ray and set the number of source views at 8 in all our experiments, we were intrigued to explore how varying the number of source views could influence the performance of the model. Therefore, we conduct an experiment with different numbers of source views on ScanNet, the results of which are presented in Table~\ref{different}. In the observation depicted in Table~\ref{different}, we can see that the utilization of extra source view images is associated with improvements in both visual and semantic segmentation quality. The improvement is more pronounced in the semantic segmentation quality as the number of source views increases. This characteristic motivates our future work to explore the design of a novel view semantic segmentation framework that operates more effectively with fewer input views.


\subsection{Compare with GeoNeRF + semhead}
In Section~\textcolor{red}{5.2.1}, we mentioned that GeoNeRF + semhead with depth supervision (second row of Table~\textcolor{red}{2}) slightly outperforms our approach (fourth row of Table~\textcolor{red}{2}) regarding PSNR and SSIM for the rendered RGB images on ScanNet, while our GSNeRF excels in all semantic segmentation metrics by approximately 5\%. To further show the advantage of our GSNeRF, we conduct a qualitative comparison in Fig.~\ref{fig:geonerf_compare}. Despite observing marginal decreases in image rendering metrics (PSNR, SSIM) compared to GeoNeRF + semhead, the perceptual impact on visual quality is not obvious. However, a more notable distinction arises in semantic segmentation quality, as GeoNeRF + semhead produces a pronounced dissimilarity in semantic content with the ground truth (denoted as GT) in Fig.~\ref{fig:geonerf_compare}.
This suggests that while GeoNeRF + semhead may marginally outperform us in image rendering metrics, our method significantly excels in delivering superior semantic segmentation results.

\subsection{More Qualitative Evaluation}
Fig.~\ref{fig:quality_image} shows more qualitative evaluation results. The first three columns of each row illustrate the novel view image synthesis results from S-Ray, our GSNeRF, and the GT image. The latter three columns present the corresponding novel view semantic segmentation outcomes for S-Ray, our proposed GSNeRF, and the GT semantic segmentation map.


\end{document}